# EXTRACTING SENTIMENT ATTITUDES FROM ANALYTICAL TEXTS

**Loukachevitch N. V.** (louk_nat@mail.ru)
Lomonosov Moscow State University, Moscow, Russia

**Rusnachenko N.** (kolyarus@yandex.ru)
Bauman Moscow State Technical University, Moscow, Russia

In this paper we present the RuSentRel corpus including analytical texts in the sphere of international relations. For each document we annotated sentiments from the author to mentioned named entities, and sentiments of relations between mentioned entities. In the current experiments, we considered the problem of extracting sentiment relations between entities for the whole documents as a three-class machine learning task. We experimented with conventional machine-learning methods (Naive Bayes, SVM, Random Forest).

# ИЗВЛЕЧЕНИЕ ОЦЕНОЧНЫХ ОТНОШЕНИЙ МЕЖДУ СУЩНОСТЯМИ ИЗ АНАЛИТИЧЕСКИХ ТЕКСТОВ

**Лукашевич Н. В.** (louk_nat@mail.ru)
МГУ имени М. В. Ломоносова, Москва, Россия

**Русначенко Н.** (kolyarus@yandex.ru)
МГТУ имени Н. Э. Баумана, Москва, Россия

## 1. Introduction

Automatic sentiment analysis, i.e. identification of opinions on the subject discussed in the text, is one of the most popular applications of natural language processing during last years.

Approaches to extracting the sentiment position from a text depends on the genre of the text being analyzed. So, one of the most studied text genres in the sentiment analysis task is users' reviews about products or services. Such texts are usually devoted to discussion on a single entity (but, perhaps in its various aspects), and the opinion is expressed by one author, namely the author of the review [Pang et al., 2002; Taboada et al., 2011; Liu 2012; Chetviorkin and Loukachevitch, 2013; Loukachevitch et al., 2015].

Another popular type of texts for sentiment analysis are short messages posted in social networks, especially, in Twitter [Pak, Paroubek, 2010; Loukachevitch, Rubtsova, 2016; Rosenthal et al., 2017]. These texts can require very precise analysis but, at the same time, they cannot contain multiple opinions toward multiple entities because of short length.

One of the most complicated genres of documents for sentiment analysis are analytical articles that analyze a situation in some domain, for example, politics or economy. These texts contain opinions conveyed by different subjects, including the author(s)' attitudes, positions of cited sources, and relations of mentioned entities to each other. Analytical texts usually contain a lot of named entities, and only a few of them are subjects or objects of a sentiment attitude. Besides, an analytical text can have a complicated discourse structure. Statements of opinion can take several sentences, or refer to an entity mentioned several sentences earlier. Also a sentence containing an opinion or describing the relationship's orientation between entities may contain other named entities, which complicates the recognition of sentiment attitudes, their subjects and objects.

In this paper, we present a corpus of analytical articles in Russian annotated with sentiments towards named entities and describe experiments for automatic recognition of sentiments between named entities. This task is a specific subtask of relation extraction.

## 2. Related Work

The task of extracting sentiments towards aspects of an entity in reviews has been studied in numerous works [Liu 2012, Loukachevitch et al., 2015]. Also extraction of sentiments to targets, stance detection was studied for short texts such as Twitter messages [Amigo et al., 2012; Loukachevitch, Rubtsova, 2016; Mohammad et al., 2017]. But the recognition of sentiments towards named entities or events including opinion holder identification from full texts has been attracted much less attention.

In 2014, the TAC evaluation conference in Knowledge Base Population (KBP) track included so-called sentiment track [Ellis et al., 2014]. The task was to find all cases where a query entity (sentiment holder) holds a positive or negative sentiment about another entity (sentiment target). Thus, this task was formulated as a query-based retrieval of entity-sentiment from relevant documents and focused only on query entities[1].

In [Deng et al., 2015], MPQA 3.0 corpus is described. In the corpus, sentiments towards entities and events are labeled. A system trained on such data should answer such questions as "Toward whom is X negative/positive?", "Who is negative/positive toward X"? The annotation is sentence-based. For example, in sentence "When the Imam issued the fatwa against Salman Rushdie for insulting the Prophet…", Imam is negative to Salman Rushdie, Salman Rushdie is negative to Prophet. Imam is also negative toward event of insulting. However, Imam is is positive toward the Prophet. The current MPQA corpus consists of 70 documents. In total, sentiments towards 4,459 targets are labeled.

The paper [Choi, et al., 2016] studied the approach to the recovery of the documents attitudes between subjects mentioned in the text. For example, from the sentence "Russia criticizes Belarus for allowing Mikhail Saakashvili appear on Belarusian TV," it is possible to infer not only the fact that Russia is dissatisfied with Belarus, but also the fact that Russia has the negative attitude toward Mikhail Saakashvili. The authors integrate several attitude constraints into the integer linear programming framework to improve attitude extraction.

To assess the quality of the approach, the text collection consisting of 914 documents was labeled. About 3 thousand opinions were found. As a result of the markup, it was found that about 25% of assessments were extracted not from a single sentence but from neighbor sentences. The best quality of opinion extraction obtained in the work was only about 36% F-measure, which shows that the necessity of improving extraction of attitudes at the document level is significant and this problem is currently studied insufficiently. Inter-annotator agreement was estimated as 0.35 for positive labels and 0.50 for negative labels (Cohen's kappa).

The inference of sentiments with multiple targets in a coherent text, additional features should be accounted for. For the analysis of these phenomena, in the works [Scheible and Schütze, 2013; Ben-Ami et al., 2014] the concept of sentiment relevance is discussed. In [Ben-Ami et al., 2014], the authors consider several types of the thematic importance of entities discussed in the text: the target—the main entity of the text; accidental—an entity only mentioned in this text; relationTarget—the theme of the text is the relation between multiple entities of the same importance; ListTarget—the text discusses several equally important entities sequentially. These types are treated differently in sentiment analysis of coherent texts.

## 3. Corpus and annotation

In order to initiate research in the field of sentiment analysis of analytical articles for the Russian language, the annotated corpus RuSentRel has been created. The source of the corpus was Internet-portal inosmi.ru, which contains, in the main, analytical articles in the domain of international politics translated into Russian from foreign languages. The collected articles contain both the author's opinion on the subject matter of the article and a large number of sentiments between the participants of the situations described in the article.

---

[1] https://tac.nist.gov/2014/KBP/Sentiment/index.html

For the documents of the assembled corpus, manual annotation of the sentiment attitudes towards mentioned named entities have been carried out. The annotation can be subdivided into two subtypes: 1) the author's relation to mentioned named entities, 2) the relation of subjects expressed as named entities to other named entities. These opinions were recorded as triples: «Subject of opinion, Object of opinion, sentiment». The sentiment position can be negative (neg) or positive (pos), for example, (*Author, USA, neg*), (*USA, Russia, neg*). Neutral opinions or lack of opinions are not recorded. In contrast to the MPQA 3.0 corpus, the sentiments are annotated for the whole documents, not for each sentence.

In some texts, there were several opinions of the different sentiment orientation of the same subject in relation to the same object. This, in particular, could be due to a comparison of the sentiment orientation of previous relations and current relations (for example, between Russia and Turkey). Or the author of the article could mention his former attitude to some subject and indicate the change of this attitude at the current time. In such cases, it was assumed that the annotator should specify exactly the current state of the relationships.

During the annotation, it became clear that it is very difficult for annotators to indicate all the relationships between named entities, because of the complexity of the texts and the large number of mentioned named entities. Therefore, the procedure of the annotation was as follows: the texts were independently labeled by two annotators, then the annotations were joined, the duplicates were deleted. Duplicates of the attitudes could additionally appear due to different names of the same object (subject), for example, the European Union and the EU. Further, the resulting annotation was checked out by a super-annotator: a small number of discrepancies were resolved, missed relationships could be added. In total, 73 large analytical texts were labeled with about 2000 relations.

To prepare documents for automatic analysis, the texts were processed by the automatic name entity recognizer, based on CRF method [Mozharova, Loukachevitch, 2016]. The program identified named entities that were categorized into four classes: Persons, Organizations, Places and Geopolitical Entities (states and capitals as states). Automatic labeling contains a few errors that have not yet been corrected. Preliminary analysis showed that the F-measure of determining the correct entity boundaries exceeds 95%, there may be some additional errors with the definition of entity types, which is auxiliary information for the sentiment analysis in the current case. In total, 15.5 thousand named entity mentions were found in the documents of the collection.

An analytical document can refer to an entity with several variants of naming (*Vladimir Putin—Putin*), synonyms (*Russia—Russian Federation*), or lemma variants generated from different word forms. Besides, annotators could use only one of possible entity's name in the describing attitudes. For correct inference of attitudes between named entities in the whole document, we provide the list of variant names for the same entity found in our corpus. The current list contains 83 sets of name variants. In such a way, we separate the sentiment analysis task from the task of named entity coreference.

The preliminary version of corpus RuSentRel was granted to the Summer school on Natural Language Processing and Data Analysis[2], organized in Moscow in 2017. The collection was divided into the training and test parts. In the current experiments we use the same division of the data. **Table 1** contains statistics of the training and test parts of the RuSentRel corpus. The last line of the table shows the average number of named entities pairs mentioned in the same sentences without indication of any sentiment to each other per a document. This number is much larger than number of positive or negative sentiments in documents, which additionally stress the complexity of the task.

**Table 1.** Statistics of RuSentRel corpus

|  | Training collection | Test collection |
|---|---|---|
| Number of documents | 44 | 29 |
| Average number of sentences per document | 74.5 | 137 |
| Average number of mentioned entities per document | 194 | 300 |
| Average number of unique named entities per document | 33.3 | 59.9 |
| Average number of positive sentiment pairs of named entities per document | 6.23 | 14.7 |
| Average number of negative sentiment pairs of named entities per document | 9.33 | 15.6 |
| Average number of neutral sentiment pairs of named entities per document | 120 | 276 |

## 4. Experiments

In the current experiment we consider the problem of extracting sentiment relations from analytical texts as a three-class supervised machine learning task. All the named entities (NE) mentioned in a document are grouped in pairs: ($NE_1$, $NE_2$), ($NE_2$, $NE_1$). All the generated pairs should be classified as having positive, negative, or neutral sentiment from the first named entity of the pair (opinion holder) to the second entity of the pair (opinion target). To support this task, we added neutral sentiments for all pairs not mentioned in the annotation and co-occurred in the same sentences into the training and test collections.

As a measure of quality of classification, we take the averaged Precision, Recall and F-measure of positive and negative classes. In the current experiments we classify only those pairs of named entities that co-occur in the same sentence at least once in a document. We use 44 documents as a training collection, and 29 documents as a test collection in the same manner as the data were provided for the Summer School mentioned in the previous section. In the current paper, we describe the application of only conventional machine learning methods: Naive Bayes, Linear SVM and Random Forest implemented in the scikit learn package[3].

---

[2]   https://miem.hse.ru/clschool/

[3]   **http://scikit-learn.org/stable/**

The features to classify the relation between two named entities according to an expressed sentiment can be subdivided into two groups. The first group of features characterizes the named entities under consideration. The second group of features describes the contexts in that the pair occurs.

The features of named entities include the following features:

- word2vec similarity between entities. We use the pre-trained model news_2015[4] [Kutuzov, Kuzmenko, 2017]. The size of the window is indicated as 20. Vectors of multiword expressions are calculated as the averaged sum of the component vectors. Using such a feature, we suppose that distributionally similar named entities (for example, from the same country) express their opinion to each other less frequently;
- the named entity type according to NER recognizer: person, organization, location, or geopolitical entity;
- the presence of a named entity in the lists of countries or their capitals. These geographical entities can be more frequent in "expressing sentiments" than other locations;
- the relative frequency of a named entity or the whole synonym group if this group is defined in a text under analysis. It is supposed that frequent named entities can be more active in expressing sentiments or can be an object of an attitude [Ben-Ami et al., 2015];.
- the order of two named entities.

It should be noted that we do not use concrete lemmas of named entities as features to avoid memorizing the relation between specific named entities from the training collection.

The second group of features describes the context in that the pair of named entities is appeared. There can be several sentences in the text where the pair of named entities occurs. Therefore each type of features includes maximal, average and minimum values of all the basic context features:

- the number of sentiment words from RuSentiLex[5] vocabulary: the number of positive words, number of negative words. RuSentiLex contains more than 12 thousand words and expressions with description of their sentiment orientation [Loukachevitch, Levchik, 2016];
- the average sentiment score of the sentence according to RuSentiLex;
- the average sentiment score before the first named entity, between named entities, and after the second named entities according to RuSentiLex;
- the distance between named entities in lemmas;
- the number of other named entities between the target pair;
- number of commas between the named entities.

Altogether, we currently utilize 54 features.

---

[4] http://rusvectores.org/

[5] http://www.labinform.ru/pub/rusentilex/index.htm

We use several baselines for the test collection: *Baseline_neg*—all pairs of named entities are labeled as negative; *Baseline_pos*—all pairs are labeled as positive, *Baseline_random*—the pairs are labeled randomly; *Baseline_distr*—the pairs are labeled randomly according to the sentiment distribution in the training collection; *Baseline_school*—the results obtained by the best team at the Summer school[6]. The results of all baselines are shown in Table 2.

**Table 2.** Baselines for sentiment extraction between named entities for RuSentRel corpus

| Baseline method | Precision | Recall | F-measure |
|---|---|---|---|
| Baseline_neg | 0.027 | 0.390 | 0.050 |
| Baseline_pos | 0.021 | 0.400 | 0.040 |
| Baseline_random | 0.039 | 0.215 | 0.065 |
| Baseline_distr | 0.045 | 0.230 | 0.075 |
| Baseline_school | 0.130 | 0.103 | 0.120 |

Table 3 shows the classification results obtained with the use of several machine learning methods. For two methods (SVM and Random Forest), the grid search of the best combination of parameters was carried out; the grid search is implemented in the same scikit-learn package. The best results were obtained with the Random Forest classifier. The parameter tuning did not improve the results, which are quite low.

**Table 3.** Results of sentiment extraction between named entities

| Method | Precision | Recall | F1 |
|---|---|---|---|
| KNN | 0.18 | 0.06 | 0.09 |
| Naïve Bayes Gauss | 0.06 | 0.15 | 0.11 |
| Naïve Bayes Bernoulli | 0.13 | 0.21 | 0.16 |
| SVM (Default values) | 0.35 | 0.15 | 0.15 |
| SVM (Grid search) | 0.09 | 0.36 | 015 |
| Random forest (Default values) | 0.44 | 0.19 | 0.27 |
| Random forest (Grid search) | 0.41 | 0.21 | 0.27 |

But we can see that the baseline results are also very low. It should be noted that the authors of the [Choi, et al., 2016], which worked with much smaller documents, reported F-measure 36%.

There is an important question about inter-annotator agreement because of the complexity of the task. In our case, the procedure is not straightforward, because we asked people to indicated only positive or negative relations between named entities but in fact they internally classified the relations into three classes including neutral. Because of the large number of the mentioned named entities in the texts, neutral relations significantly prevail.

---

[6]  https://miem.hse.ru/clschool/results

We used the following approach. We asked another super-annotator (see **Section 3**) to label the collection, and compared her annotation with our gold standard using average F-measure of positive and negative classes in the same way as for automatic approaches. In such a way, we can reveal the upper border for automatic algorithms. We obtained that F-measure of human labeling is 0.55. This is quite low value, but it is significantly higher than the results obtained by automatic approaches. About 1% of direct contradictions (positive vs. negative) among all etalon labels were found.

## 5. Analysis of errors

In this section we consider several examples of erroneous classification of relations between entities.

1) In the following example, the system did not detect that Liuhto is positive towards NATO. This is because of relatively long distance between Liuhto and NATO and absence of evident sentiment words.

   *Лиухто говорит, что он начал склоняться к вступлению Финляндии в НАТО.*
   (*Liuhto says that he began to welcome Finland's accession to NATO.*)

2) In the following sentence, the evident sentiment words are also absent, and the system misses the sentiment from Putin to Russia,

   *Путин хочет войти в историю как царь, расширивший территорию России.*
   (*Putin wants to go down in history as the king who expanded the territory of Russia.*)

3) In the following text fragment, there are several sentences about the stance of Sven Mikser towards Putin. But in the first sentence, his position is expressed too complex. The relation is discussed also in the next sentences but pronoun resolution is needed:

   *Глава комиссии по иностранным делам эстонского парламента Рийгикогу, бывший министр иностранных дел Свен Миксер (Sven Mikser) считает, что, возможно, президент Владимир Путин не стремится присоединить к России в первую очередь страны Балтии, но подобные намерения вполне могут существовать.*

   *The head of the Foreign Affairs Committee of the Estonian Parliament, the Riigikogu, the former Foreign Minister Sven Mikser believes that, perhaps, President Vladimir Putin does not seek to join the Baltic countries first of all, but such intentions may well exist.*

We can see the large number of long distances between entities having positive or negative attitudes to each other. So, our next step is experiment with convolutional neural networks, which can represent such word sequences using multiple filters.

To enhance our training collection semi-automatically, we plan to gather sentences describing known relations, for example, Russia—Ukraine, or United Stated—Bashar Asad. But most such relations are negative.

## Conclusion

In this paper we presented the RuSentRel corpus including analytical texts in the sphere of international relations. For each document, we annotated sentiments from the author to mentioned named entities, and sentiments of relations between mentioned entities.

In the current experiments, we considered the problem of extracting sentiment relations between entities for the whole documents as a three-class machine learning task. We experimented with conventional machine-learning methods (Naive Bayes, SVM, Random Forest). The corpus RuSentRel is published[7].

Our next step is experiments with convolutional neural networks, which can represent long word sequences using multiple filters. We plan to enhance our training collection semi-automatically, trying to find sentences describing known relations, to obtain enough data for training neural networks.

## Acknowledgments

This work is partially supported by RFBR grant № 16-29-09606.

---

[7] https://github.com/nicolay-r/RuSentRel